%% file: main.tex
\documentclass[sigconf]{acmart}

\usepackage{algorithm} 
\usepackage{algorithmic} 
\usepackage{subfig}
\usepackage{float}
\AtBeginDocument{%
  \providecommand\BibTeX{{%
    \normalfont B\kern-0.5em{\scshape i\kern-0.25em b}\kern-0.8em\TeX}}}

\copyrightyear{2023}
\acmYear{2023}
\setcopyright{acmlicensed}\acmConference[KDD '23]{Proceedings of the 29th ACM SIGKDD Conference on Knowledge Discovery and Data Mining}{August 6--10, 2023}{Long Beach, CA, USA}
\acmBooktitle{Proceedings of the 29th ACM SIGKDD Conference on Knowledge Discovery and Data Mining (KDD '23), August 6--10, 2023, Long Beach, CA, USA}
\acmPrice{15.00}
\acmDOI{10.1145/3580305.3599481}
\acmISBN{979-8-4007-0103-0/23/08}

\begin{document}

\title{Quantitatively Measuring and Contrastively Exploring Heterogeneity for Domain Generalization}
\author{Yunze Tong}
\orcid{0009-0005-0305-0059}
\email{tyz01@zju.edu.cn}
\affiliation{%
  \institution{Zhejiang University}
  \city{Hangzhou}
  \country{China}
}

\author{Junkun Yuan}
\orcid{0000-0003-0012-7397}
\email{yuanjk@zju.edu.cn}
\affiliation{%
  \institution{Zhejiang University}
  \city{Hangzhou}
  \country{China}
}

\author{Min Zhang}
\orcid{0009-0001-1289-1571}
\email{zhangmin.milab@zju.edu.cn}
\affiliation{%
  \institution{Zhejiang University}
  \city{Hangzhou}
  \country{China}
}

\author{Didi Zhu}
\orcid{0009-0004-6892-5357}
\email{didi_zhu@zju.edu.cn}
\affiliation{%
  \institution{Zhejiang University}
  \city{Hangzhou}
  \country{China}
}

\author{Keli Zhang}
\orcid{0000-0002-7883-0552}
\email{zhangkeli1@huawei.com}
\affiliation{%
  \institution{Noah's Ark Lab, Huawei Technologies}
  \city{Shenzhen}
  \country{China}
}

\author{Fei Wu}
\orcid{0000-0003-2139-8807}
\email{wufei@zju.edu.cn}
\affiliation{%
  \institution{Zhejiang University}
  \city{Hangzhou}
  \country{China}
}

\author{Kun Kuang}
\orcid{0000-0001-7024-9790}
\authornote{Corresponding author.}
\email{kunkuang@zju.edu.cn}
\affiliation{%
  \institution{Zhejiang University}
  \city{Hangzhou}
  \country{China}
}

\renewcommand{\shortauthors}{Yunze Tong et al.}

\input{parts/abstract.tex}



\maketitle

\input{parts/introduction.tex}

\input{parts/related-work.tex}

\input{parts/problem_setting.tex}

\input{parts/preliminary.tex}

\input{parts/method.tex}

\input{parts/experiment.tex}

\input{parts/conclusion.tex}

\bibliographystyle{ACM-Reference-Format}
\bibliography{reference}

\input{parts/appendix}

\end{document}

%% file: parts/abstract.tex
\begin{abstract}  
  Domain generalization (DG) is a prevalent problem in real-world applications, which aims to train well-generalized models for unseen target domains by utilizing several source domains. Since domain labels, \textit{i.e.}, which domain each data point is sampled from, naturally exist, most DG algorithms treat them as a kind of supervision information to improve the generalization performance. However, the original domain labels may not be the optimal supervision signal due to the lack of domain heterogeneity, \textit{i.e.}, the diversity among domains. For example, a sample in one domain may be closer to another domain, its original label thus can be the noise to disturb the generalization learning. Although some methods try to solve it by re-dividing domains and applying the newly generated dividing pattern, the pattern they choose may not be the most heterogeneous due to the lack of the metric for heterogeneity. In this paper, we point out that domain heterogeneity mainly lies in variant features under the invariant learning framework. With contrastive learning, we propose a learning potential-guided metric for domain heterogeneity by promoting learning variant features. Then we notice the differences between seeking variance-based heterogeneity and training invariance-based generalizable model. We thus propose a novel method called \textbf{H}eterogeneity-based \textbf{T}wo-stage \textbf{C}ontrastive \textbf{L}earning (HTCL) for the DG task. In the first stage, we generate the most heterogeneous dividing pattern with our contrastive metric. In the second stage, we employ an invariance-aimed contrastive learning by re-building pairs with the stable relation hinted by domains and classes, which better utilizes generated domain labels for generalization learning. Extensive experiments show HTCL better digs heterogeneity and yields great generalization performance.
\end{abstract}

%% file: parts/introduction.tex
\vspace{-10pt}
\section{Introduction}\label{section: introduction}
Machine learning has exhibited extraordinary and super-human performance on various tasks \cite{resnet, Bert}. 
Its common paradigm, Empirical Risk Minimization (ERM) \cite{vapnik1999overview}, trains models by minimizing the learning error on training data under the independent and identically distributed (i.i.d.) assumption.
However, distribution shifts, \textit{i.e.}, out-of-distribution (OOD) problems,  usually exist between training and real-world testing data, which leads to the performance degradation of ERM-based methods~\cite{geirhos2020shortcut, engstrom2020identifying, recht2019imagenet, ref:lv2023duet}. 
In order to solve the OOD generalization problem, researchers turn to developing methods for domain generalization (DG), a task that aims to generalize the models learned in multiple source domains to other unseen target domains. 
In DG, the features of samples can be divided into two parts: invariant (\textit{a.k.a.} semantic or stable) features and variant (\textit{a.k.a.} non-semantic or unstable) features. 
Invariant features are those having a domain-invariant relation with true class labels, while variant features shift with the change of domains. Thus, DG methods tend to perform invariant learning: identify the stable relations among all source domains and predict class labels with invariant parts. 
ERM-based approaches may easily learn and rely on variant features for prediction \cite{geirhos2020shortcut} because they even do not tell which domain the training samples are from at all. Assuming that all data come from one distribution, they achieve high accuracies in seen domains but might degrade a lot in unseen target domains. 

\begin{figure}[t]
  \centering
  \includegraphics[width=0.45\textwidth]{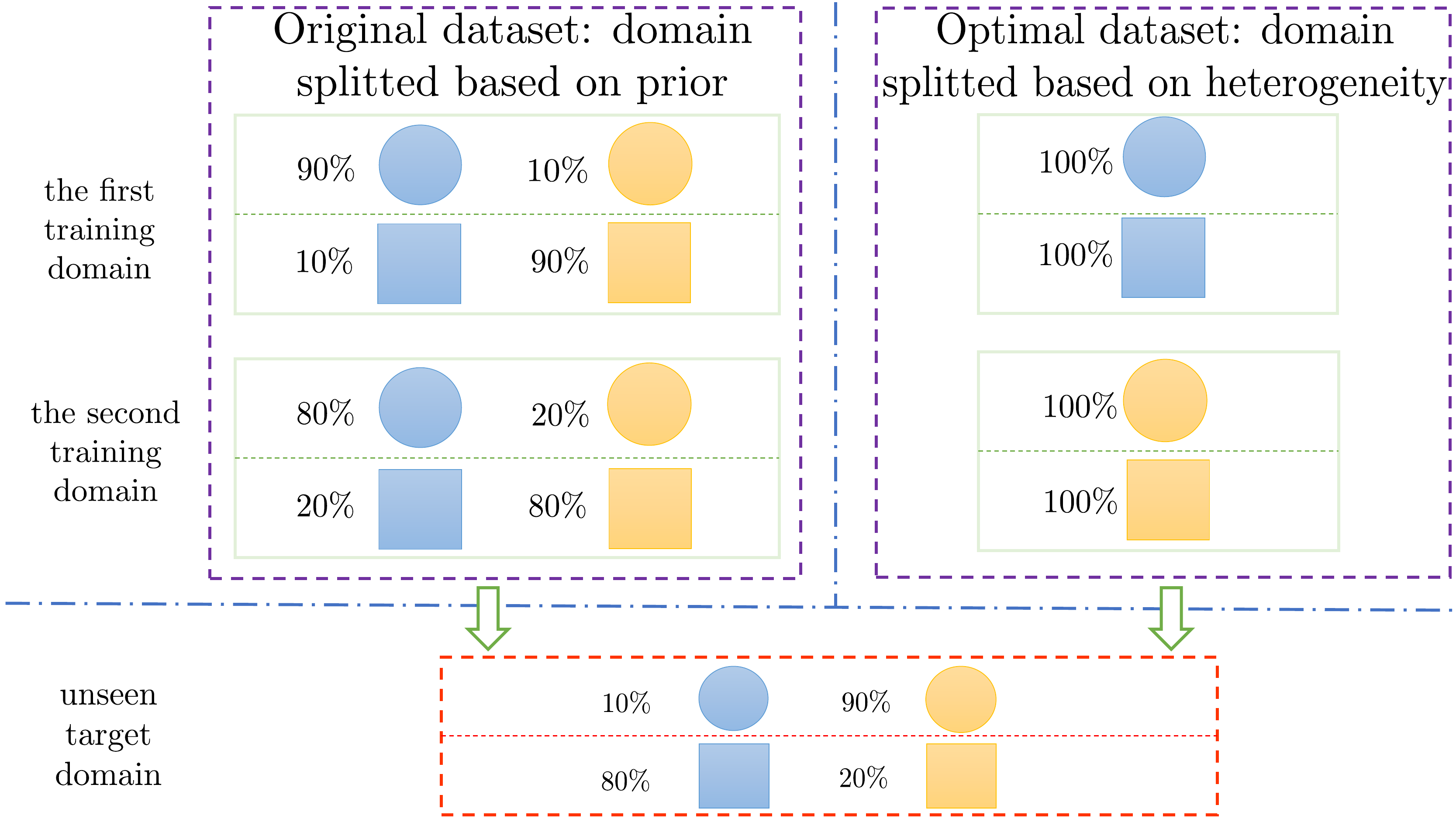}
  \vspace{-10pt}
  \caption{The toy example for illustrating the problem of lack in heterogeneity. See main text for details.}\label{img: toy example}
  \vspace{-20pt}
\end{figure}

To tackle this issue, some methods turn to using the gap between different domains to surpass ERM. They utilize the original domain labels as extra supervision signals \cite{IRM, Sagawa2020GroupDRO, blanchard2021mtl,lv2023ideal, lv2022personalizing}. Data are separately treated according to which domain they are sampled from. 
Among them, some fine-grained methods try to minimize some statistical metrics (\textit{e.g.}, MMD \cite{CORAL, li2018domain} or Wasserstein distance \cite{zhou2020domain}) among feature representations of different domains. Models trained by these methods are more robust to complex domains than ERM and show good performances in bridging the domain gap. However, these methods neglect the quality of the original domain labels they use, \textit{i.e.}, the risk of lacking heterogeneity. We illustrate this phenomenon with Figure~\ref{img: toy example}. Here we aim to predict classes by shapes, which are invariant features, while the colors are domain-variant. In realistic scenarios, we may follow a make-sense prior that making the samples within a domain diverse as the left of Figure~\ref{img: toy example}. However, when performing DG, the left pattern will easily make the model predict labels with domain-variant features because their domains own two very similar distributions in colors. In fact, the optimal pattern is shown on the right of Figure~\ref{img: toy example}, which makes the domain-variant features heterogeneous across domains and homogeneous within each domain. As illustrated above, sub-optimal domain labels will introduce bias during training and mislead the generalization learning.
To reduce the bias, enlarging domain heterogeneity, \textit{i.e.}, the variety across domains, is an effective approach. 
From this view, domain labels could be regarded as heterogeneity-focused dividing according to prior. 
If source domains are heterogeneous enough, models may be more robust to unseen domains because source sub-populations provide more kinds of data distribution during training. 
However, if source domains are homogeneous, the latent data distribution and predictive strategy within every source domain may tend to be similar. 
Models may tend to rely on a common predictive path for training, which is unstable for unseen distributions. 
Therefore, digging domain heterogeneity is significant for reducing bias in DG. 

 Recently, some methods notice the importance of data heterogeneity and turn to utilize heterogeneity to boost generalization learning. A more heterogeneous dataset is believed to be able to help disentangle features better and cut off the shortcut of neural networks to variant features \cite{EIIL}.
They mostly generate a dividing pattern (\textit{a.k.a.} infer environments or split groups) to re-separate training samples into newly generated domains. By applying their patterns, they achieve favorable accuracies \cite{EIIL, HRM, IP-IRM}. However, there is no precise definition and metric for heterogeneity in the DG task yet. Without the metric during domain label generation, the chosen dividing pattern may be sub-optimal, which might even bring new noise and disturb generalization learning. In addition, their experiments are mainly based on synthetic or low-dimension data, which is insufficient to verify their potential in dividing domains and generalization abilities in real-world scenarios.

In this paper, we propose a quantitative learning potential-guided heterogeneity metric and introduce a heterogeneity-based two-stage DG algorithm through contrastive learning. 
We first point out that domain heterogeneity mainly lies in variant features in DG under the invariant learning framework. 
Our metric is calculated with the ratio of the average distance between the representations of different domains, which needs to construct and contrast the same-class pairs within each domain and across domains. When it involves measuring the heterogeneity of a pattern, we apply the metric to the features from a variance-focused model, which also indicates the potential to obtain heterogeneity. 
Our method comprises generating heterogeneous patterns and enhancing generalization with the pattern. We select the most heterogeneous dividing pattern from the generated ones in the first stage, which is measured with our heterogeneity metric quantitatively. The first stage contains two interactive modules: the heterogeneity exploration module and the pattern generation module. They are performed iteratively and boosted by each other. In the second stage, with the domain labels generated by the first stage, we construct positive pairs with same-class and different-domain data while negative ones indicate different-class data within the same domain. 
Then an invariance-aimed contrastive learning is employed to help train a well-generalized model for the DG task.

To summarize, our contributions include:
\begin{itemize}
    \item We point out the heterogeneity in DG, \textit{i.e.}, the original domain labels may not be optimal when treated as supervision signals. Sub-optimal domain labels will introduce bias and disturb the generalization learning.
    \item Under the insight that heterogeneity mainly lies in variant parts, we propose a metric for measuring domain heterogeneity quantitatively. Through the features from a variance-focused model, our metric contrast the same-class
pairs within each domain and across domains.
    \item We propose a heterogeneity-based two-stage contrastive learning method for DG. It incorporates both generating heterogeneous dividing patterns and utilizing generated domain labels to guide generalization learning.
    \item Extensive experiments exhibit the state-of-the-art performance of our method. Ablation studies also demonstrate the significance of each stage in our method.
\end{itemize}

%% file: parts/related-work.tex
\section{Related Work}\label{section: related work}
We first provide an overview of the related work involved with our setting and method.

{\bfseries Domain generalization.} Domain generalization (DG) aims to train a well-generalized model which can avoid the performance degradation in unseen domains \cite{yuan2023domain}. Domains are given as natural input in DG, which means models can access more information. However, sub-optimal signals will also induce distribution shift \cite{MixStyle} or make the model rely on spurious features \cite{IRM}. That's why DG may be more hard to solve compared to ordinary classification tasks. Several different approaches are proposed to tackle this problem. Some works use data augmentation to enlarge source data space \cite{fourier4IL, adversarial_data_augmentation, MixStyle, yuan2022label, wu2021indirect, zhang2022tree}. Methods based on GroupDRO \cite{Sagawa2020GroupDRO} or IRM \cite{IRM} turn to optimize one or some specific groups' risk to achieve higher generalization \cite{lin2022zin, zhou2022sparse, lin2022bayesian}. 
There are also some methods employing feature alignment \cite{Fish, bai2021decaug,zhu2023universal} or decorrelation \cite{liao2022decorr} to guide the learning procedure. 

{\bfseries Heterogeneity in domain generalization.} Data heterogeneity broadly exists in various fields. In DG, heterogeneity mainly refers to the diversity of each domain. It is caused by the distribution shifts of data when dividing the implicit whole data distribution into different sub-populations, \textit{i.e.}, domains. 
Though there are some formulations for data heterogeneity in other machine learning problems \cite{anonymous2023measure, li2020federated, HTE}, domain heterogeneity has no precise and uniform metric till now. EIIL \cite{EIIL} is one of the first methods that notice the gain brought by re-dividing domains in DG. It infers domain labels with a biased model. HRM \cite{HRM} designs a framework where dividing domains and performing invariant learning are optimized jointly. KerHRM \cite{KerHRM} develops HRM by integrating the procedure with kernel methods to better capture features. IP-IRM \cite{IP-IRM} also takes the strategy of dividing domains to better disentangle features from a group-theoretic view. Above novel methods achieve favorable performances on synthetic or low-dimension data. However, they all treat generating dividing patterns as a middle process and haven't proposed the metric for domain heterogeneity. In addition, their potentials in high-dimension data, which are less decomposed on the raw feature level, are not fully verified. These are the issues we aim to make up in this paper.


{\bfseries Invariant learning.} Invariant learning actually can be seen as one of the effective approaches in DG. Its core is differentiating invariant and variant parts in an image and making the model tend to learn the invariant features across different domains. Following this line, some works \cite{MatchDG, CIRL, yuan2023instrumental, miao2022domain, ziyu2023differentiated} achieve the goal by the pre-defined causal graph to learn the key features involving class labels. Some works also consider the disentangled method \cite{CSD, montero2020role, li2023multi, zhang2022towards, niu2023knowledge, zhou2022model, 9712445} aiming to split the features completely. Besides, ZIN \cite{lin2022zin} proposes to use explicit auxiliary information to help learn stable and invariant features, which is also a promising direction.

{\bfseries Contrastive learning.} Contrastive learning (CL) contrasts semantically similar and dissimilar pairs of samples, which aims to map positive sample pairs closer while pushing apart negative ones in the feature space. 
The prototype of CL is the architecture of Contrastive Predictive Coding (CPC)  \cite{oord2018representation}. It contains InfoNCE loss which can be optimized to maximize a lower bound on mutual information in the theoretical guarantee. 
Contrastive learning has already achieved great success in self-supervised learning \cite{chen2020simple, he2020momentum} due to the need of modeling unlabeled data. 
It has been applied in various tasks due to the ability to make the hidden representations of the samples from the same class close to each other \cite{li2022hero, zhang2021causerec, li2023winner, zhang2023fairnessaware}.

%% file: parts/problem_setting.tex
\section{Problem Setting}\label{problem setting}
We formalize the problem setting in domain generalization (DG) task. Suppose that we have source data $\mathcal{D} = \mathcal{X} \times \mathcal{Y} $ for a classification task, every sample $x_i$ follows $x_i \in \mathcal{X}$ and its label $y_i$ follows $y_i \in \mathcal{Y}$. Typically, we need a non-linear feature extractor $\phi$ and a classifier $w$. $\phi$ will map the sample from input space to representation space, which can be formalized as $\phi: \mathcal{X} \rightarrow \mathcal{H}$. $\mathcal{H}$ denotes the representation space here. Then $\phi(x_i) \in \mathbb{R}^d$ denotes the $d$-dimension feature representation of $x_i$. $w(\phi(x_i)) \in \mathbb{R}^{|\mathcal{Y}|}$ denotes the predicted possibilities for all labels. In DG, every sample has its initial domain label. Let $\varepsilon^{tr}$ be the training domains (\textit{a.k.a.} environments or groups). Then we use $d_{x_i} \in \varepsilon^{tr}$ to denote the domain label of the training sample $x_i$. 

Some solutions don't rely on domain labels. For example, Empirical Risk Minimization (ERM) \cite{vapnik1999overview} minimizes the loss over all the samples no matter which domain label they have:
\begin{equation}
    \mathcal{L}_{ERM} = E_{(x_i, y_i) \in \mathcal{D}}[\ell(w(\phi(x_i)), y_i)],
\end{equation}
where $\ell$ is the Cross Entropy loss function for classification. For some other methods which utilize domain labels, the risk will have an extra penalty term. Take Invariant Risk Minimization (IRM) \cite{IRM} as an example, 
\begin{equation}\label{eq: IRM}
    \mathcal{L}_{IRM} = \sum\limits_{\epsilon \in \varepsilon^{tr}} \mathit{R}^{\epsilon}(\phi, w) + \lambda\left\| \nabla_{\bar{w}} \mathit{R}^{\epsilon}(\bar{w} \circ \phi)\right\|.
\end{equation}
Here $\mathit{R}^{\epsilon}(\phi, w) = E_{\left\{(x_i, y_i)|d_{x_i} = \epsilon \right\}} [\ell(w(\phi(x_i)), y_i)]$ denotes the per-domain classification loss. The second term in Equation (\ref{eq: IRM}) enforces simultaneous optimality of the $\phi$ and $w$ in every domain. $\bar{w}$ is a constant scalar multiplier of 1.0 for each output dimension.

No matter whether to use domain labels, the aims of DG methods are consistent: training a robust model which can generalize to unseen domains with only several source domains.

%% file: parts/preliminary.tex
\section{Domain Heterogeneity}\label{section: preliminary}

\subsection{Variance-based Heterogeneity}\label{section: example for motivation}
In domain generalization (DG), all the samples are always given with their specific domain labels due to the need of differentiating source domains during training. Therefore, domain labels reflect the process of splitting data into several sub-populations, which denotes domain heterogeneity. Domain heterogeneity has a close connection with training models. Suppose that all the training and testing data form an implicit overall distribution. After splitting them into several domains, the data points in each domain will form a complete data distribution and have their specific predictive mechanisms. If the source domains are homogeneous, the latent distribution and predictive strategy within every source domain may tend to be similar. Models trained in these domains will also tend to learn a common predictive mechanism because they can hardly access novel distributions. Therefore, when facing the target unseen domains, the robustness of the model may not be favorable. Therefore, pursuing domain heterogeneity is important for DG.

Here we explain where domain heterogeneity lies. Following the idea of invariant learning, the feature representation obtained from supervised learning is composed of invariant features and variant features. The invariant parts will help predict labels while the variant parts may harm training. Therefore, common methods use techniques to make the model sensitive to the invariant parts in latent space as much as possible instead of the variant ones. 
For a pair of same-label samples, their invariant parts should be naturally close because they own the same class label, which highly involves invariant features. While their variant parts can show arbitrary relations. For example, two samples whose labels are both dogs may have similar invariant parts like dogs' noses. But as for variant parts, they can be various.
They can be similar if they both have grass as the background while they can be totally different if one is in the grass and the other is in the water.\footnote{In some DG datasets, the variant parts mainly refers to the style of the whole picture instead of the background in our example.} An ideal heterogeneous dividing pattern focuses more on the \textbf{variant} parts. The reason is that: the target unseen domain will share the same invariant information (because they are highly involved with the label information, which is shared among every domain) with the source domains, while the variant parts from the target domain may be unseen during training. Thus, if we make the variant parts of source domains homogeneous within a single domain and heterogeneous among different domains, we will maximally imitate the situation of facing unseen domains during training. 
Therefore, our goal comes to: \textbf{re-divide domain labels for every sample to form a more heterogeneous dividing pattern than original one}.

\subsection{The Metric of Heterogeneity}\label{section: heterogeneity metric}
Since the heterogeneity among source domains counts, how to measure it quantitatively becomes the main problem. 
Obviously, different dividing patterns (different allocation of domain labels for all samples) have different heterogeneity because of the bias in the variant parts. 
Suppose that we let a model with fixed architecture learn the representation with a dividing pattern. 
If the dividing pattern is consistent with the assumptions in section \ref{section: example for motivation}, \textit{i.e.}, the variant parts are homogeneous enough within each domain and heterogeneous enough across different domains, then the following requirement should be fulfilled: for any pair of same-class samples, the distance between their representations should be as small as possible if they are from the same domain while the distance should be as large as possible if they are from different domains.

Considering that the heterogeneity lies in the variant features as stated above, we characterize the heterogeneity quantitatively with the proportion between the distances of different groups of the same-class features:
\begin{equation}
    \mathcal{L}_H = \sum\limits_{(x, y)} log(\frac
    {E_{\{(x', y')|d_{x'} = d_{x}, y' = y\}}[||\phi(x) - \phi(x')||_2]}
    {E_{\{(x'', y'')|d_{x''} \neq d_{x}, y'' = y\}}[||\phi(x) - \phi(x'')||_2]}).
    \label{formula: metric for heterogeneity}
\end{equation}
For a sample $(x, y)$, we collect all its same-class sample ($y=y'=y''$). Then we set $x$'s positive (negative) pair samples $x'$ ($x''$) as the ones which are from the same (different) domain $d_{x'} = d_{x}$ ($d_{x''} \neq d_{x}$).
The more heterogeneous the dividing pattern is, the less will the numerators become and the more will the denominators become. $\mathcal{L}_H$ thus will become less. We use Equation (\ref{formula: metric for heterogeneity}) to judge whether the current dividing pattern is more heterogeneous in the iterative process.
We calculate the distance between the features instead of the original data here. Recall section \ref{section: example for motivation}, we can have such intuition because we can differentiate the variant or invariant part just with original images. But as for a fixed model, the representation it learns will lose information because of the reduction of the dimensions. From the view of neural networks, it is the similarity between low-dimension representation mainly works. As a result, if we want to measure heterogeneity, we should concentrate on the similarity between full-trained representations instead of the original data. Therefore, we put stress on the potential for learning heterogeneous features with the given dividing pattern. We then use an already-trained model $\phi$ to generate features instead of using other methods (\textit{e.g.}, kernel tricks) only performing on the data itself.

Then it comes to how to train $\phi$. Note that our metric for heterogeneity stresses the potential to learn the best variant representation in a given dividing pattern. Therefore, training variance-based $\phi$ should minimize the loss containing classification error and the guidance for exploring heterogeneity, \textit{i.e.}, variant features in DG:
\begin{equation}
     \mathcal{L}_{var} = \sum\limits_{(x, y)}\ell(w(\phi(x)), y) + \lambda_1\mathcal{L}_H.
     \label{formula: loss for heterogeneity}
\end{equation}
The first term $\ell$ makes sure $\phi$ focus on the feature which carries helpful information for classification (no matter if it is an invariant feature or variant feature). The second term is our metric for heterogeneity. Minimizing it promotes the variant features becoming different across different domains and becoming similar in the same domain. Hyperparameter $\lambda_1$ balances the two terms.

\subsection{The Utilization of Heterogeneity}\label{section: utilize herterogeneity}
Given an ideal heterogeneous dividing pattern following our metric, it comes to training feature extractor $\phi$ and classifier $w$ for the classification task. Here we aim to make feature extractor $\phi$ focus more on the invariant parts in each image. 
With the reorganized domain labels from the last stage, the variant parts of the samples from the same domain would be homogeneous.
Therefore, we transfer our target to other two kinds of sample pairs: the sample pairs with the same domain label and different class labels, and the sample pairs with different domain labels and the same class label. It is obvious that in representation space, the features should follow the strategy of making intra-class distance small and inter-class distance large.
We continue using the dog's example in section \ref{section: example for motivation} to explain the motivation. Here we have already obtained a heterogeneous dividing pattern by reorganizing domain labels. The model should learn the invariant features (\textit{e.g.}, animals' noses or hairs) which will help distinguish whether the sample is a dog or cat, not the variant features (\textit{e.g.}, the background or the style of the image) which tend to be homogeneous within a single domain and heterogeneous across different domains. Suppose that there are two domains that are heterogeneous, the first domain has images of cats or dogs in the grass and the second domain has same-label images whose background is water. $\phi$ fulfilling $dist(\phi(x_{dog, grass}), \phi(x_{dog, water})) \ll dist(\phi(x_{dog, grass}), \phi(x_{cat, grass}))$ would be more appropriate for predicting the label. To achieve the goal of encoding more invariant features rather than variant features, we leverage two specific kinds of sample pairs mentioned above.
For $(x,y) \in \mathcal{D}$, its negative pair samples are images having different class labels from $y$ and the same domain label as $d_x$:
\begin{equation*}
    P_x^{-} = \{x'|d_{x'}=d_x, y' \neq y\}.
\end{equation*}
The positive pair samples are images having the same class label as $y$ and different domain labels from $d_x$:
\begin{equation*}
    P_x^{+} = \{x'|d_{x'} \neq d_x, y' = y\}.
\end{equation*}
We use Maximum Mean Discrepancy (MMD) as an estimation of the similarity of a pair of features. Then we write a penalty term, which contrasts the pairs to promote utilizing domain heterogeneity, as:
\begin{small}
\begin{equation}\label{formula: penalty term for prediction}
    -\sum\limits_{x}log\frac
    {\sum\limits_{x_- \in P_x^{-}}mmd(\phi(x), \phi(x_-))}
    {\sum\limits_{x_- \in P_x^{-}}mmd(\phi(x), \phi(x_-)) + \sum\limits_{x_+\in P_x^{+}}mmd(\phi(x), \phi(x_+))}.
\end{equation}
\end{small}

Here we compare the metric in Equation (\ref{formula: metric for heterogeneity}) and the regularization term in Equation (\ref{formula: penalty term for prediction}) again.
Their aims are totally different. In Equation (\ref{formula: metric for heterogeneity}), the core lies in utilizing the relations among variant features, which in turn helps train $\phi$ for digging the potential of enlarging heterogeneity in Equation (\ref{formula: loss for heterogeneity}). However, when it involves utilizing heterogeneity, the ultimate goal is training the best feature extractor $\phi$ and classifier $w$ for classification. 
In DG, the ideal $\phi$ will only encode invariant features and not encode variant features to representation space at all. So our aim is to make the feature extractor focus more on the invariant features.

%% file: parts/method.tex
\section{Proposed Methods}
Our heterogeneity-based two-stage contrastive learning (HTCL) method is based on the problem setting in section~\ref{problem setting} and the metric in section~\ref{section: heterogeneity metric}. 
Following the analysis above, we treat the procedure of exploring heterogeneity and the one of training models with newly generated domain labels separately. The first stage aims to generate a heterogeneous dividing pattern. It can be seen as a preprocessing stage that only re-divides domain labels and outputs the most heterogeneous ones, which corresponds to section \ref{section: heterogeneity metric}. 
The second stage receives the domain labels from the first stage as input. During its training process, it utilizes the contrastive loss mentioned in section \ref{section: utilize herterogeneity} to help learn invariant features. We detail these two stages in section \ref{section: generate pattern} and \ref{section: prediction}. Note that the two stages have no iteration process. 
They are performed only once in a consistent order while the modules \textbf{within} the first stage have an iterative process. 
That's a difference from existing methods \cite{EIIL, IP-IRM, HRM, KerHRM} mentioned in section~\ref{section: introduction}, which also aim to generate domain labels by themselves but incorporate dividing domains and invariant learning together. Algorithm~(\ref{pseudo code}) shows the whole process of HTCL.

\begin{figure}[h]
  \centering
  \includegraphics[width=\linewidth]{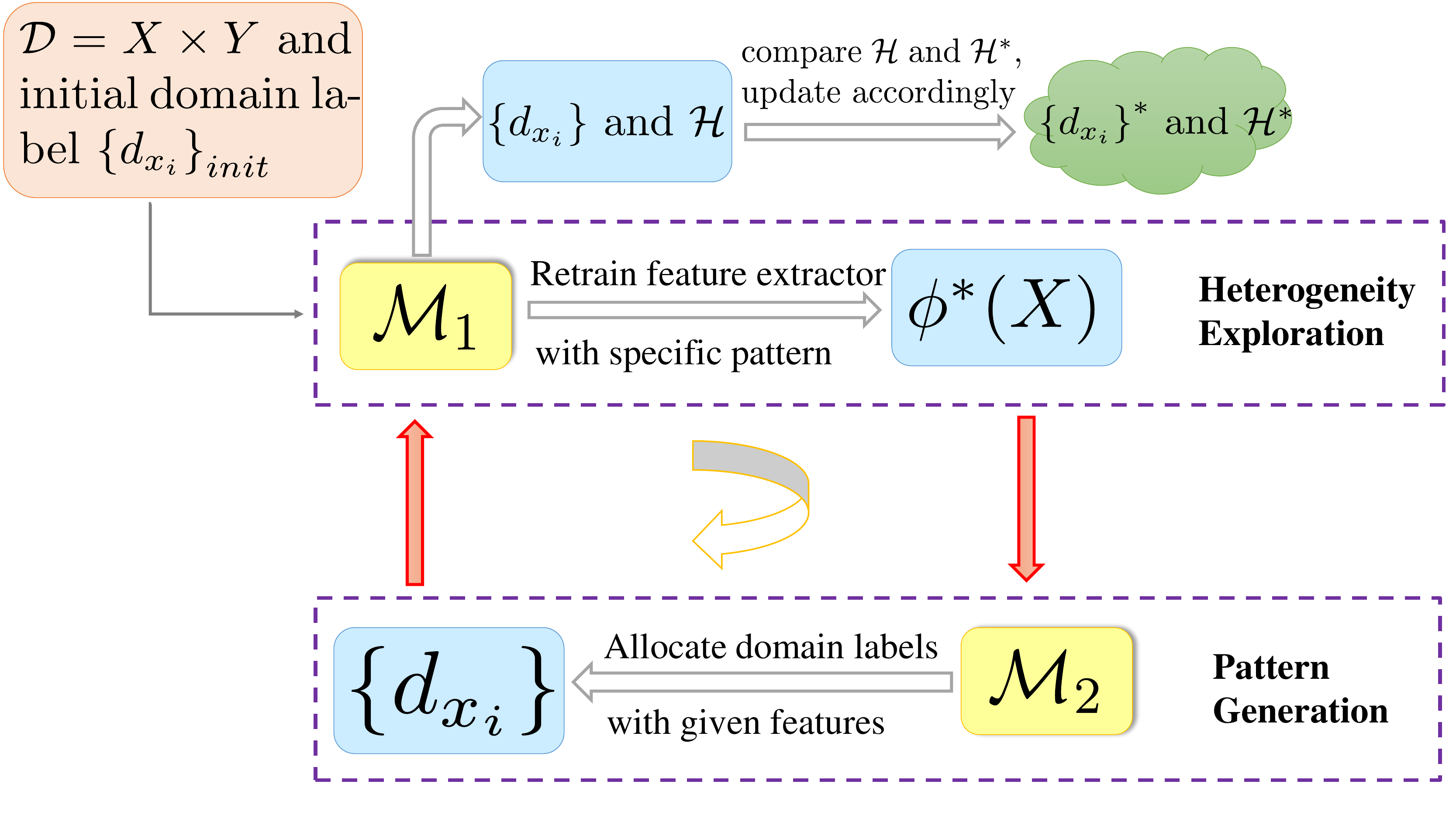}
  \vspace{-17pt}
  \caption{The framework for the first stage of HTCL. $\mathcal{M}_1$ denotes the heterogeneity exploration module, and $\mathcal{M}_2$ refers to the pattern generation module. $\mathcal{M}_1$ outputs a well-trained feature representation for $\mathcal{M}_2$. $\mathcal{M}_2$ returns a newly generated dividing pattern to $\mathcal{M}_1$. The input and output of each module are shown in blue boxes. The final optimal output $\left\{d_{x_i}\right\}^*$, which will be used as the domain labels in ultimate invariant learning, is shown in green box. It updates according to the heterogeneity metric during the iterations. }\label{img: framework}
  \Description{The whole framework of our method to dig an optimal heterogeneous dividing pattern.}
  \vspace{-17pt}
\end{figure}

\subsection{Heterogeneous Dividing Pattern Generation}\label{section: generate pattern}
With the training set and original domain labels, we explore heterogeneity by dividing images into $|\varepsilon^{tr}|$ disjoint domains. However, exploring heterogeneity with variant features and measuring heterogeneity can't be optimized simultaneously. As a result, we design two interactive modules and make them work alternately.

\subsubsection{Heterogeneity exploration module}\label{section: feature heterolyze}
A fixed dividing pattern and raw samples are given as input. Our goal includes:
\begin{itemize}
    \item learning an adequate $\phi$ guided by Equation (\ref{formula: loss for heterogeneity}) and generating the low-dimension representation $\sum\limits_{x_i} \phi(x_i)$ of all samples for the next pattern generation module.
    \item measuring the heterogeneity of the input dividing pattern with the metric $\mathcal{L}_H$ quantitatively.
\end{itemize}

For the first goal, we can just update $\phi$ and $w$ through Equation~(\ref{formula: loss for heterogeneity}). 
When it refers to constructing the corresponding positive and negative pairs for $\mathcal{L}_{var}$, we only construct them within every batch instead of the whole dataset. 

Here we re-write $\mathcal{L}_H$ which has been mentioned in Equation~(\ref{formula: metric for heterogeneity}). $\mathcal{L}_H$ is our heterogeneity metric, and we also apply it in Equation~(\ref{formula: loss for heterogeneity}) to guide heterogeneity exploration. Considering the calculation is conducted per batch, we re-formalize $\mathcal{L}_H$ more detailedly as: 
\begin{equation}
    \mathcal{L}_H = - \sum\limits_{(X_i, y_i)}\sum\limits_{(d_m, d_n)} \log\frac{dist(\phi(X_i^{d_m}), \phi(X_i^{d_n}))}{dist(\phi(X_i^{d_m})) + dist(\phi(X_i^{d_n}))} .
\end{equation}
We denote a single batch by $B = \{(x, y)^{|B|}\} \subset \mathcal{D}$. For each kind of label $y_i \in \mathcal{Y}$, $X_i$ denotes the set of all images which own class label $y_i$ in the batch $B$. $d_{X_i}$ denotes the set of all domains which contain at least one sample in $X_i$. $\forall (d_m, d_n)\ \mathrm{s.t.}\ d_m \in d_{X_i}, d_n \in d_{X_i}, d_m \neq d_n$, $X_i^{d_m}$ and $X_i^{d_n}$ denotes the samples whose domain labels are $d_m$ and $d_n$ separately. For convenience, here we set the input of feature extractor $\phi$ as a set of images and every dimension of the features denotes the predicted probability of each label for a single image. Then we set function $dist(\cdot, \cdot)$ to measure the average distance between different kinds of representation:
\begin{equation}
    dist(\phi_1, \phi_2) = \frac{\sum\limits_{i=0}^{|\phi_1|-1}\sum\limits_{j=0}^{|\phi_2|-1}\left\|\phi_1[i] - \phi_2[j]\right\|_2}{|\phi_1|\times|\phi_2|}.
\end{equation}
In above equation, $\phi_1$ has a shape of $|\phi_1| \times d$ and the entry $\phi_1[i]$ denotes the representation of $i$-th image in $\phi_1$. $\phi_2$ has a similar form. We then calculate the average distance between different representations. Similarly, we can measure the average distance among same-domain and same-label samples as follows:
\begin{equation}
    dist(\phi) = \frac{\sum\limits_{i=0}^{|\phi|-2}\sum\limits_{j=i+1}^{|\phi| - 1}\left\|\phi[i] - \phi[j]\right\|_2}{|\phi|^2 - |\phi|}.
\end{equation}
After training for several epochs, we obtain the representation with the final $\phi^*$ as the optimal feature extractor. Then it comes to the second goal: measuring the heterogeneity. Similarly, we calculate the ratio batchwisely and finally take the expectation among all batches' counted pairs to avoid the problem brought by batches' random shuffle:
\begin{equation}\label{eq: measure heterogeneity batchwisely}
    H = E_{B \subset \mathcal{D}}[\mathcal{L}_{H}].
\end{equation}
Thus, $H$ specifies the quantity of the heterogeneity in a given dividing pattern. Note that when performing Equation (\ref{eq: measure heterogeneity batchwisely}), we use a trained feature extractor $\phi^*$ to replace $\phi$ in $\mathcal{L}_{H}$, which makes sure Equation (\ref{eq: measure heterogeneity batchwisely}) reflect the maximum potential of the variance-based model in such a dividing pattern.

\noindent\textbf{Remark.}\textit{ When it refers to heterogeneity, the subjective is feature representation rather than raw data itself. A fully-trained representation reflects the learning potential of a fixed model with a given dividing pattern. So we choose to train $\phi$ in such above Equation (\ref{formula: loss for heterogeneity}) manner before measuring heterogeneity instead of directly performing clustering methods (\textit{e.g.}, density-based) on raw data.}

\input{parts/pseudocode.tex}

\subsubsection{Pattern generation module}\label{section: generate pattern innerly}

In this module, input $\phi^*(\mathcal{X})$ is the low-dimension representation of all the samples from the last module. We divide them into $|\varepsilon^{tr}|$ disjoint domains to form a new dividing pattern. We use a simple multilayer perceptron (MLP) $f(\cdot)$ to map representation to $|\varepsilon^{tr}|$ domains with probability: $f(\phi^*(\mathcal{X})) \in \mathbb{R}^{|\mathcal{X}|\times|\varepsilon^{tr}|}$. No supervised signals are used in this module. But we should avoid two situations:

\textbf{Generating process is influenced by invariant features too much.} Considering that heterogeneity exploration module inevitably uses class-label as supervision information, the input of this module will naturally carry invariant features. Suppose that $\phi^*$ can mainly extract invariant features of samples $x_1, x_2, x_3$ in the early stage. If $x_1$ and $x_2$ share the same class while $x_3$ doesn't, the distance between $\phi^*(x_1)$ and $\phi^*(x_2)$ will naturally be less than the distance between $\phi^*(x_1)$ and $\phi^*(x_3)$ no matter whether $x_1$ and $x_3$ share the similar variant features or not. In a word, same-label samples tend to be close to each other. Then they may be divided into one domain when invariant features dominate a main position in the representation space. There are two drawbacks. 
On the one hand, when feeding this kind of dividing pattern into heterogeneity exploration module, we can't construct effective pairs to boost the process of learning variant features and measuring heterogeneity. 
On the other hand, if domain labels are highly involved with class labels, the existence of domain labels will be of no help to the final process of learning semantic invariant features. Commonly, the number of domains is less than the number of labels\footnote{Actually, all the datasets we used in the experiment fulfill this condition}
. From the view of information theory, the information brought by domain labels is the subset of the information brought by class labels. Then our final training process will degrade to ERM. Therefore, we should avoid this module generating dividing patterns by actual classes. 

\textbf{The numbers of samples contained in each domain differ a lot.} If a domain contains few samples while another domain contains a lot, it will cause similar problems mentioned in the last point. Thus, we should also balance the numbers in all the domains.

The overall predictive probability of mapped domains, \textit{i.e.}, $f(\phi^*(\mathcal{X}))$, has a shape of $\mathbb{R}^{|\mathcal{X}|\times|\varepsilon^{tr}|}$. We calculate along the first dimension to obtain all samples' average probability distribution: $f(\phi^*(\mathcal{X}))_{avg} \in \mathbb{R}^{1\times|\varepsilon^{tr}|}$. Then we guide $f(\cdot)$ which generates candidate domain labels as follows:
\begin{equation}\label{eq: generate new dividing pattern}
    \mathcal{L}_{divide} = H(f(\phi^*(\mathcal{X}))_{avg}) + (\frac{1}{|\epsilon^{tr}|} - \min(f(\phi^*(\mathcal{X}))_{avg})).
\end{equation}
Note that $\phi^*$ is fixed here and we only update $f(\cdot)$. $H(p)$ denotes the entropy of the distribution $p$. The first term uses entropy on classes to encourage the same-label samples not to shrink to a single domain, which reduces the influence of invariant features. The second term is a penalty term to avoid the minimum sample number in one domain becoming too small.

\subsubsection{Summary of generating a heterogeneous dividing pattern}
Generating a heterogeneous dividing pattern is the core stage in our method. We design two interactive modules to iteratively learn the optimal dividing pattern which contains more variant information. At the start, we use original domain labels as the initial dividing pattern and send them to the heterogeneity exploration module. Then we update $\phi$ with Equation (\ref{formula: loss for heterogeneity}) and measure the heterogeneity with Equation (\ref{eq: measure heterogeneity batchwisely}). After obtaining the optimal feature representation, we send it into the pattern generation module to generate a new dividing pattern with Equation (\ref{eq: generate new dividing pattern}). The newly generated dividing pattern is then sent as input to the heterogeneity exploration module again. We repeat this procedure several times to learn variant features as possible and select the best dividing pattern which obtains the minimum value in Equation (\ref{eq: measure heterogeneity batchwisely}). Figure \ref{img: framework} illustrates the whole framework.

\input{tables/main_table.tex}
\subsection{Prediction with Heterogeneous Dividing Pattern}\label{section: prediction}

In this part, our aim changes from learning a heterogeneous dividing pattern to learning invariant features from the samples to help predict the labels in the unseen target domain. Apart from the standard classification loss, we add a distance-based term to prevent the representations between different domains too far. 
We measure the distance between two different representations as:
\begin{equation}\label{eq: mmd}
    mmd(D_{1}, D_{2}) = \frac{1}{4d^2}\left\| Cov(D_{1}) - Cov(D_{2}) \right\|_{F}^{2}.
\end{equation}
$Cov(\cdot)$ means covariances and $\left\|\cdot\right\|_{F}^{2}$ denotes the squared matrix Frobenius norm. $d$ still denotes the number of dimensions of the feature representation. The invariance-based contrastive loss for better utilizing the heterogeneous domain labels takes the form as: 
\begin{equation}
    \mathcal{L}_{cont} = \sum\limits_{\epsilon\in\varepsilon^{tr}}\sum\limits_{y\in\mathcal{Y}}\log(1+\frac{mmd(\phi(X_{s}), \phi(X_{pos}))}{mmd(\phi(X_{s}), \phi(X_{neg}))}).
\end{equation}
Consistent to section \ref{section: utilize herterogeneity}, here $X_{s}$ denotes all the source samples belonging to domain $\epsilon$ and owning label $y$. $X_{pos}$ contains all the postive pair samples for $X_{s}$, which are not from domain $\epsilon$ but own the same class as $y$. $X_{neg}$ are from $\epsilon$ while have different classes from $y$. 
Then we calculate all the distances between different domain representations like the method in CORAL \cite{CORAL}:
\begin{equation}
    \mathcal{L}_{mmd} = \sum\limits_{\epsilon_1, \epsilon_2\in\varepsilon^{tr},\  \epsilon_1 \neq \epsilon_2} mmd(\phi(X_{\epsilon_1}), \phi(X_{\epsilon_2})).
\end{equation}
Then the ultimate objective function for DG turns to:
\begin{equation}\label{eq: predictive loss}
    \mathcal{L}_{predict} = \sum\limits_{(x,y)}\ell(w(\phi(x)), y) + \lambda_{cont}\mathcal{L}_{cont} + \lambda_{mmd}\mathcal{L}_{mmd}.
\end{equation}
 $\lambda_{cont}$ and $\lambda_{mmd}$ are the hyperparameters to balance each guidance.

%% file: parts/pseudocode.tex
\begin{algorithm}[b]
    \renewcommand{\algorithmicrequire}{\textbf{Input:}}
	\renewcommand{\algorithmicensure}{\textbf{Output:}}
    \caption{The pseudo code of HTCL} 
	\label{pseudo code} 
	\begin{algorithmic}
        \REQUIRE training samples $\mathcal{D} = \mathcal{X} \times \mathcal{Y}$, original domain labels $\left\{ d_{x_i}\right\}_{init}$, hyperparamters $\lambda_1, \lambda_{cont}, \lambda_{mmd}, T_1, T_2$
        \STATE \textbf{// The first stage: heterogeneous dividing pattern generation}
		\STATE Initialization: $\phi, w$ initialized with pretrained ResNet, $\left\{ d_{x_i}\right\} \leftarrow \left\{ d_{x_i}\right\}_{init}$, $H^* \leftarrow 0$
        \REPEAT
		\STATE Update $\phi$ based on Equation (\ref{formula: loss for heterogeneity}) till convergence 
        \STATE $\phi^* \leftarrow \phi$
        \STATE Measure heterogeneity of $\left\{ d_{x_i}\right\}$ with Equation (\ref{eq: measure heterogeneity batchwisely}), obtain $H$ 
        \IF{$H < H^* \vee H^* == 0$}
        \STATE $H^* = H$
        \STATE $\left\{ d_{x_i}\right\}^* = \left\{ d_{x_i}\right\}$
        \ENDIF
        \STATE Generate a new dividing pattern $\left\{ d_{x_i}\right\}$ with Equation (\ref{eq: generate new dividing pattern}) 
		\UNTIL reach $T_1$ iterations
        \STATE \textbf{// The second stage: prediction with heterogeneous dividing pattern}
        \STATE Initialization: $\phi, w$ initialized with pretrained ResNet again, $\left\{ d_{x_i}\right\}^*$ received from the first stage
		\REPEAT
        \STATE Update $\phi, w$ with Equation (\ref{eq: predictive loss}) using $\left\{ d_{x_i}\right\}^*$ as domain labels
        \UNTIL reach $T_2$ iterations
		\ENSURE $\phi, w$
	\end{algorithmic}  
\end{algorithm}

%% file: tables/main_table.tex
\begin{table*}[t]
    \caption{The comparison with other DG methods in standard DomainBed \cite{gulrajani2021in} benchmark.  We report average accuracy on all target domains under three runs. The results of baselines are from their original corresponding literatures or DomainBed. We highlight the best results with boldface.}
    \label{table: full_table}
    \vspace{-8pt}
    \begin{tabular}{lcccc|c}
    \toprule
    \textbf{Algorithm}        & PACS & VLCS             & OfficeHome       & TerraIncognita     & \textbf{Avg.}              \\
    \midrule
    ERM \cite{vapnik1999overview}                & 85.5\scriptsize$\pm0.2$          & 77.5\scriptsize$\pm0.4$          & 66.5\scriptsize$\pm0.3$                & 46.1\scriptsize$\pm1.8$                             & 68.9           \\
    IRM \cite{IRM}                & 83.5\scriptsize$\pm0.8$          & 78.5\scriptsize$\pm0.5$          & 64.3\scriptsize$\pm2.2$                & 47.6\scriptsize$\pm0.8$                             & 68.4           \\
    GroupDRO \cite{Sagawa2020GroupDRO}           & 84.4\scriptsize$\pm0.8$          & 76.7\scriptsize$\pm0.6$          & 66.0\scriptsize$\pm0.7$                & 43.2\scriptsize$\pm1.1$                             & 67.5           \\
    Mixup \cite{yan2020mixup}              & 84.6\scriptsize$\pm0.6$          & 77.4\scriptsize$\pm0.6$          & 68.1\scriptsize$\pm0.3$                & 47.9\scriptsize$\pm0.8$                             & 69.5           \\
    MLDG \cite{li2018mldg}               & 84.9\scriptsize$\pm1.0$          & 77.2\scriptsize$\pm0.4$          & 66.8\scriptsize$\pm0.6$                & 47.7\scriptsize$\pm0.9$                             & 69.2           \\
    CORAL \cite{CORAL}              & 86.2\scriptsize$\pm0.3$          & \textbf{78.8}\scriptsize$\pm0.6$          & 68.7\scriptsize$\pm0.3$                & 47.6\scriptsize$\pm1.0$                             & 70.4           \\
    MMD \cite{li2018domain}                & 84.6\scriptsize$\pm0.5$          & 77.5\scriptsize$\pm0.9$          & 66.3\scriptsize$\pm0.1$                & 42.2\scriptsize$\pm1.6$                             & 67.7           \\
    DANN \cite{ganin2016dann}               & 83.7\scriptsize$\pm0.4$          & 78.6\scriptsize$\pm0.4$          & 65.9\scriptsize$\pm0.6$                & 46.7\scriptsize$\pm0.5$                             & 68.7           \\
    CDANN \cite{li2018cdann}              & 82.6\scriptsize$\pm0.9$          & 77.5\scriptsize$\pm0.1$          & 65.7\scriptsize$\pm1.3$                & 45.8\scriptsize$\pm1.6$                             & 67.9           \\
    MTL \cite{blanchard2021mtl}                & 84.6\scriptsize$\pm0.5$          & 77.2\scriptsize$\pm0.4$          & 66.4\scriptsize$\pm0.5$                & 45.6\scriptsize$\pm1.2$                             & 68.5           \\
    SagNet \cite{nam2019sagnet}             & 86.3\scriptsize$\pm0.2$          & 77.8\scriptsize$\pm0.5$          & 68.1\scriptsize$\pm0.1$                & 48.6\scriptsize$\pm1.0$                             & 70.2           \\
    ARM \cite{zhang2020arm}                & 85.1\scriptsize$\pm0.4$          & 77.6\scriptsize$\pm0.3$          & 64.8\scriptsize$\pm0.3$                & 45.5\scriptsize$\pm0.3$                             & 68.3           \\
    VREx \cite{krueger2020vrex}               & 84.9\scriptsize$\pm0.6$          & 78.3\scriptsize$\pm0.2$          & 66.4\scriptsize$\pm0.6$                & 46.4\scriptsize$\pm0.6$                             & 69.0           \\
    RSC \cite{huang2020rsc}                & 85.2\scriptsize$\pm0.9$          & 77.1\scriptsize$\pm0.5$          & 65.5\scriptsize$\pm0.9$                & 46.6\scriptsize$\pm1.0$                             & 68.6           \\
    CAD \cite{ruan2021CAD}      & 85.2\scriptsize$\pm0.9$         & 78.0\scriptsize$\pm0.5$         & 67.4\scriptsize$\pm0.2$               & 47.3\scriptsize$\pm2.2$                            & 69.5           \\
    CausalRL-CORAL \cite{chevalley2022causalRL}     & 85.8\scriptsize$\pm0.1$          & 77.5\scriptsize$\pm0.6$          & 68.6\scriptsize$\pm0.3$                & 47.3\scriptsize$\pm0.8$                             & 69.8           \\
    CausalRL-MMD \cite{chevalley2022causalRL}       & 84.0\scriptsize$\pm0.8$          & 77.6\scriptsize$\pm0.4$          & 65.7\scriptsize$\pm0.6$                & 46.3\scriptsize$\pm0.9$                             & 68.4           \\    
    \midrule
    HTCL (Ours)               & \textbf{88.6}\scriptsize$\pm0.3$ & 77.6\scriptsize$\pm0.5$         & \textbf{71.3}\scriptsize$\pm0.6$       & \textbf{50.9}\scriptsize$\pm1.9$             & \textbf{72.1}         \\ 
    \bottomrule
    \end{tabular}
\end{table*}

%% file: parts/experiment.tex
\section{Experiments}

\subsection{Experimental Settings}

We first introduce the common settings of our experiments. 

\noindent\textbf{Datasets.}
We use four kinds of domain generalization (DG) datasets to evaluate our proposed HTCL method:
\begin{itemize}
    \item PACS \cite{PACS} contains 9991 images shared by 7 classes and 4 domains \{art, cartoon, photo, and sketch\}.
    \item OfficeHome \cite{OfficeHome} contains 15588 images shared by 65 classes and 4 domains \{art, clipart, product, real\}.
    \item VLCS \cite{VLCS} contains 10729 images shared by 5 classes and 4 domains \{VOC2007, LabelMe, Caltech101, SUN09\}.
    \item TerraIncognita \cite{TerraIncognita} comprises photos of wild animals taken by
cameras at different locations. Following \cite{gulrajani2021in}, we use 4 domains \{L100, L38, L43, L46\}, which contains 24330 images shared by 10 classes.
\end{itemize}

These four datasets show different kinds of shift in DG. PACS and OfficeHome are mainly distinguished by images' style by human eyes, while VLCS and TerraIncognita divide domains by different backgrounds, which involves spurious correlation with actual labels. Note that our method stresses exploring the heterogeneity in representation space. The method therefore can be fit for both two kinds of shift because representation is needed for both learning processes, which gains an advantage over only style-based or only causality-based DG methods.

\noindent\textbf{Evaluation metric.}
We use the training and evaluation protocol presented by DomainBed benchmark \cite{gulrajani2021in}. In DG, a domain is chosen as unseen target domain and other domains can be seen as source domains for training. Following the instruction of the benchmark, we split each source domain into 8:2 training/validation splits and integrate the validation subsets of each source domain to create an overall validation set, which is used for validation. The ultimate chosen model is tested on the unseen target domain, and we record the mean and standard deviation of out-of-domain classification accuracies from three different runs with different train-validation splits. For one dataset, we set its every domain as test domain once to record the average accuracy and integrate the prediction accuracy of every domain by a second averaging to stand for the performance.  

\noindent\textbf{Implementation details.}
Our method comprises two stages: generating a heterogeneous dividing pattern and training a prediction model. 
In the first phase, we use ResNet-18 \cite{resnet} pre-trained on ImageNet \cite{imagenet} as the backbone feature extractor $\phi$. We change ResNet-18's last FC layer's output to a low 64-dimension for saving computation time. We additionally set a classifier whose input dimension is 64 to help classification. For generating new dividing patterns, we use a multilayer perceptron (MLP) to divide samples into domains. The MLP has 3 hidden layers with 256 hidden units. Finally, we send the optimal domain labels for all the training samples to the ultimate training stage and the networks trained in heterogeneity generation stage are dropped. As for the hyper-paramters referred in Algorithm~\ref{pseudo code}, we set $T_1=5, \lambda_1=0.01, \lambda_{cont}=1, \lambda_{mmd}=1$ from HTCL. The value of $T_2$ follows the default value in DomainBed. When training the ultimate model for predicting, we follow DomainBed's strategy. We use ResNet-50 \cite{resnet} pre-trained on ImageNet \cite{imagenet} as the backbone network for all datasets, and we use an additional FC layer to map features to classes as a classifier. As for model selection, we turn to SWAD \cite{SWAD} for weight averaging.

\subsection{Main Results}
We first demonstrate our results and compare HTCL with various domain generalization methods in Table \ref{table: full_table}. Our method achieves the best performance on most of the datasets. The performance of our method exceeds that of CORAL \cite{CORAL} by 1.7\% on average. Specifically, the performance gain on TerraIncognita, which is the hardest to predict among all these four dataset, is delightful, reaching 2.3\% over SagNet \cite{nam2019sagnet} method. All the above comparisons reveal the effect of our method and further demonstrate the improvement brought by seeking and utilizing heterogeneity.

\input{tables/ablation.tex}

\begin{figure}[t]
  \centering
  \subfloat[EIIL]{\includegraphics[width=\linewidth]{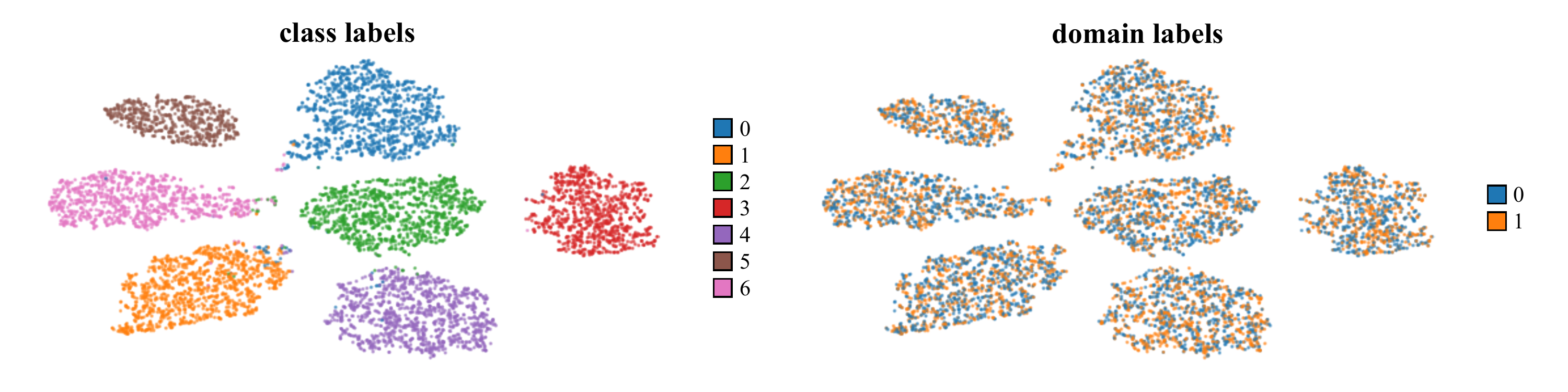}}\\
  \subfloat[KerHRM]{\includegraphics[width=\linewidth]{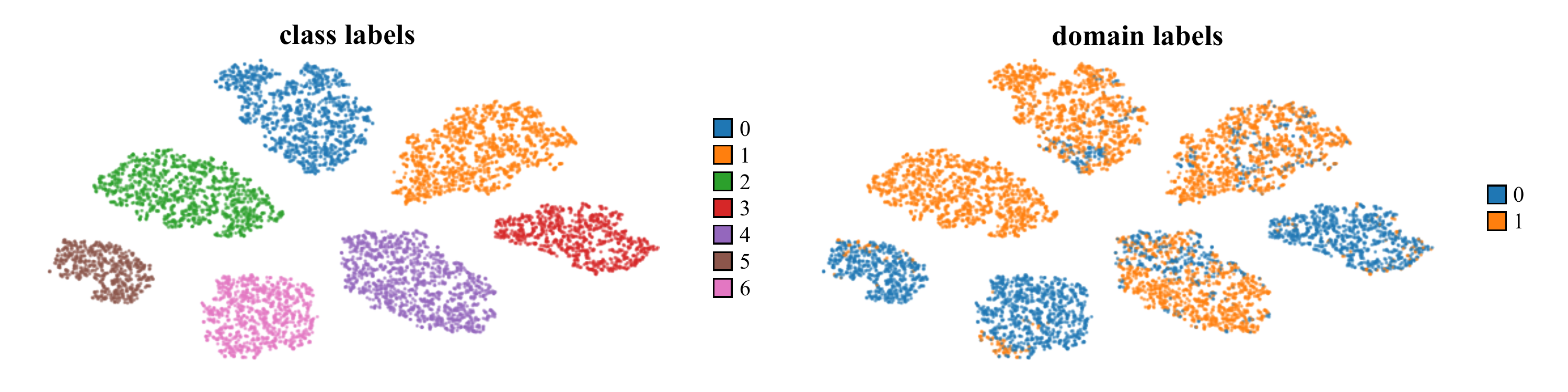}}\\
  \subfloat[HTCL (Ours)]{\includegraphics[width=\linewidth]{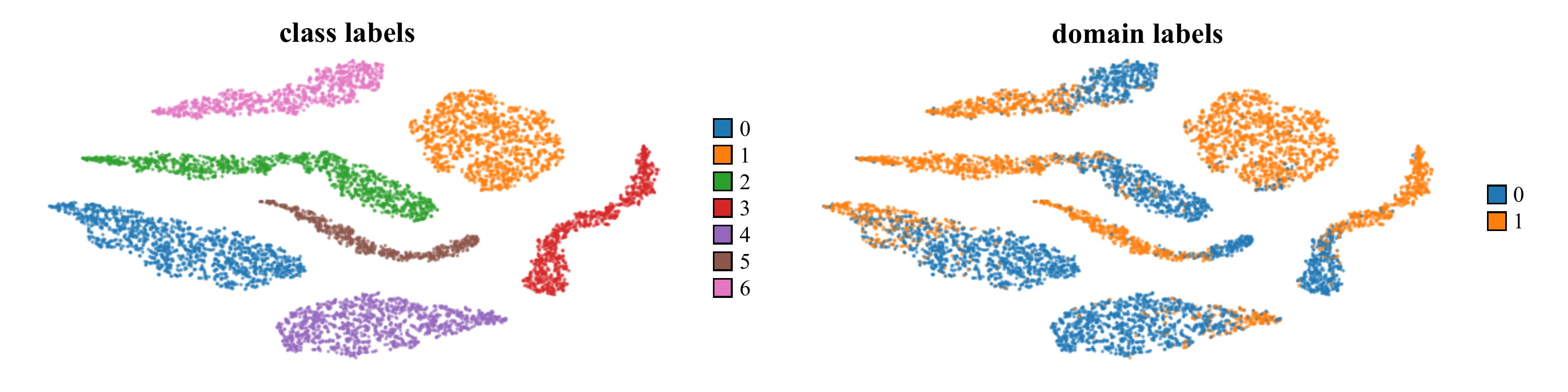}}
  \vspace{-10pt}
  \caption{T-SNE \cite{t-SNE} of the feature representations during generating domain labels in three methods. All these three methods generate dividing patterns with the help of learned feature representations. We obtain the features and compare samples' relative distances with t-SNE. The colors of left column reflect classes while the ones of right denote domains.}
  \label{img: ablation comparison}
  \Description{Samples' t-sne with domains generated by three methods}
  \vspace{-15pt}
\end{figure}

\subsection{Comparison with Other Similar Methods}
Our method aims for utilizing heterogeneity to help prediction and we implement this goal by generating a heterogeneous dividing pattern, \textit{i.e.}, domain labels. This strategy of re-allocating domain labels is shared by several previous methods. Thus, we compare the performances of these methods under the DomainBed framework. These previous methods have no official version for the DG datasets we used and they mainly report their performances on low-dimension data. So we re-implement them with the help of their code and integrate them into the benchmark framework. 
As for the hyperparameters, we mainly follow their recommendation and finetune some of them for better performances. 
The comparison results on the PACS dataset, which is one of the most fashioned DG datasets, are shown in Table \ref{tab: comparison among similar methods}. We report both their performances on target seen domains (In-domain) and performances on target unseen domains (Out-of-domain). We can find that our method is more suitable for DG tasks than other previous methods. All these methods enjoy a rather high In-domain accuracy. However, as for out-of-domain performances, our method shows better performances than KerHRM \cite{KerHRM} by 4.7\%. In addition, our method's standard deviation on testing accuracies is also lower than other methods, which indicates the robustness of HTCL. The results confirm the problem mentioned in section \ref{section: introduction}: though the novel ideas of these methods are effective in their experiments on synthetic or low-dimension data, the representation in DG datasets is more complex and hard to disentangle. Therefore, some tricks in the feature level will degrade in bigger DG datasets. That's why we design our method with the help of supervision information to obtain a better representation for classification as mentioned in section \ref{section: heterogeneity metric}.

To further explore these methods' process of dividing domain labels, we use t-SNE \cite{t-SNE} to visualize the feature representations of three methods: EIIL, KerHRM, and our HTCL. All these three methods generate new dividing patterns based on the learned features in an interactive module.
In other words, the features they learn will influence the process of domain label generation. We obtain the features which decide the final domain labels in these methods and assess their quality. Figure \ref{img: ablation comparison} illustrates the scattering results of their features with class labels and domain labels separately by different colors. 
It can be seen that the features of the three methods all form clusters with class labels. 
While as for the right figures which are differentiated with domain labels, it can be seen that EIIL tends to separate the same-class samples into different domains too fairly. The domains split by KerHRM follow the strategy to lower intra-domain distances and enlarge inter-domain distances better than EIIL. However, the same-class samples may tend to be divided into one domain in KerHRM, which has the risk of class imbalance among domains. In our method, the data within every manifold is divided into both domains more or less, which indicates the role of our guidance (Equation \ref{eq: generate new dividing pattern}) in generating candidate patterns.

\input{tables/comparison_result.tex}

\subsection{Ablation Study}
Considering that the process of generating dividing patterns has no interaction with ultimate training, it is necessary to split the phases to evaluate our methods. We conduct ablation studies to investigate every module's role in helping enhance generalization. We consider three factors: totally dropping the first stage that generates heterogeneous dividing patterns, replacing the candidate pattern generating guidance (Equation (\ref{eq: generate new dividing pattern})) with simple K-Means cluster algorithm \cite{K-Means} in the first stage, and dropping the contrastive term in the ultimate training stage which aims for utilizing heterogeneity. The results of performances on PACS are listed in Table \ref{tab: ablation}. It confirms that every lack of the module will lower the generalization ability of the model compared to the original HTCL method.

\subsubsection{Effects of applying heterogeneous dividing pattern.} We first totally drop the stage of re-allocating domain labels. Instead, we use the original domain labels, which turns into the common setting of DomainBed. As shown in Table \ref{tab: ablation}, dropping the procedure of digging heterogeneity lowers the predictive accuracy by 0.4\%, which shows the significance of creating domain-heterogeneous dataset in DG. 

\subsubsection{Effects of our guidance to generate specific candidate patterns.} Generating candidate dividing patterns is necessary for learning variant features. In pattern generation module, we design Equation~(\ref{eq: generate new dividing pattern}) to guide the split of domains. Here we replace this module with a common K-Means cluster algorithm on the given low-dimension representation. As seen in Table \ref{tab: ablation}, our original pattern generation method outperforms the K-Means method by 0.4\%. In fact, simply using the K-means algorithm to generate candidate dividing patterns has similar results with totally dropping the first module (both of them obtain 88.2\% testing accuracy on PACS).

\subsubsection{Effects of the contrastive term for utilizing heterogeneity.} As for the ultimate training, we add an extra contrastive term in the loss function to improve model's generalization ability. We conduct the experiments without this term. As shown in Table \ref{tab: ablation}, 
the average accuracy reduces from 88.6\% to 88.3\% when dropping this term.

\subsubsection{Summary of the ablation study.} Above ablation study confirms that every part of HTCL plays a role in helping generalization. There is another observation. In PACS, when setting target unseen domain as C (cartoon) and S (sketch), models' testing accuracies are worse than setting that as A (art) and P (photo), which means samples from C or S are more hard to perform prediction when being set as target domain. Obviously, enhancing the testing accuracies of these domains is more valuable. We note that maintaining the original HTCL framework outperforms in these domains, which indicates our method achieve robustness by pursuing heterogeneity.

%% file: tables/ablation.tex
\begin{table*}[t]
\caption{Summary of the ablation study on PACS. The second row block contains the ablation on the modules of the first stage (generating heterogeneous dividing pattern), while the third row block denotes the ablation on the second stage (predicting with heterogeneous dividing pattern). We record the degradation compared with original method in the last column.}
\label{tab: ablation}
\vspace{-8pt}
\begin{tabular}{lcccc|c} 
\toprule
 & A & C & P & S & Avg. ($\Delta$)  \\ \midrule
without generating heterogeneous dividing pattern         & 89.4 \scriptsize$\pm0.3$ & 83.0 \scriptsize$\pm0.5$ & 97.7 \scriptsize$\pm0.1$ & 82.9 \scriptsize$\pm1.0$ & 88.2 (-0.4) \\ 
replace the guidance of pattern generation with K-Means & 90.3 \scriptsize$\pm1.0$ & 82.6 \scriptsize$\pm1.0$ & 97.0 \scriptsize$\pm0.3$ & 83.1 \scriptsize$\pm0.7$ & 88.2 (-0.4) \\ 
\midrule
without predicting with heterogeneous dividing pattern       & 90.1 \scriptsize$\pm0.4$ & 82.5 \scriptsize$\pm0.2$ & 97.3 \scriptsize$\pm0.2$ & 83.2 \scriptsize$\pm1.3$ & 88.3 (-0.3) \\
\midrule
HTCL (Original)           & 89.7 \scriptsize$\pm0.6$ & 82.9 \scriptsize$\pm0.4$ & 97.5 \scriptsize$\pm0.5$ & 84.1 \scriptsize$\pm0.4$ & 88.6  \\
\bottomrule
\end{tabular}
\end{table*}

%% file: tables/comparison_result.tex
\vspace{-8pt}
\begin{table}[hbp]
\caption{The comparison results among the methods which also generate new dividing pattern for DG on PACS \cite{PACS} dataset. In-domain results denote the testing performance on the target seen domains. Out-of-domain results denote that on the target unseen domain. The standard deviation of testing runs is reported after the average testing accuracies.}
\vspace{-8pt}
\label{tab: comparison among similar methods}
\begin{tabular}{lcc} 
\toprule
                & Out-of-domain & In-domain  \\
\midrule
EIIL \cite{EIIL}           & 81.7 \scriptsize$\pm0.6$ & 93.2 \scriptsize$\pm0.3$  \\
IP-IRM \cite{IP-IRM}         & 81.7 \scriptsize$\pm0.4$ & 97.1 \scriptsize$\pm0.6$  \\
KerHRM \cite{KerHRM}         & 83.9 \scriptsize$\pm2.3$ & 97.5 \scriptsize$\pm0.1$  \\
\midrule
HTCL (Ours)            & \textbf{88.6} \scriptsize$\pm0.3$ & \textbf{97.8} \scriptsize$\pm0.1$  \\
\bottomrule
\end{tabular}
\end{table}
\vspace{-17pt}

%% file: parts/conclusion.tex
\section{Conclusion}
In this work, we comprehensively consider the role of domain labels in the domain generalization (DG) task and explain why domain-heterogeneous datasets can help model obtain better performances on unseen domains. By building a connection between variant features and heterogeneity, we propose a metric for measuring heterogeneity quantitatively with contrastive learning. Besides, we notice that an invariance-aimed contrastive learning performed in the ultimate training will make model better utilize the information brought by heterogeneous domains. Thus, we integrate both digging heterogeneity and utilizing heterogeneity in one framework to help train well-generalized models. We denote this integrated method as heterogeneity-based two-stage contrastive learning (HTCL) for DG. Extensive experimental results show the effectiveness of HTCL on complex DG datasets.

\begin{acks}
This work was supported in part by National Natural Science Foundation of China (62006207, 62037001, U20A20387), Young Elite Scientists Sponsorship Program by CAST (2021QNRC001), Zhejiang Province Natural Science Foundation (LQ21F020020), Project by Shanghai AI Laboratory (P22KS00111), Program of Zhejiang Province Science and Technology (2022C01044), the StarryNight Science Fund of Zhejiang University Shanghai Institute for Advanced Study (SN-ZJU-SIAS-0010), and the Fundamental Research Funds for the Central Universities (226-2022-00142, 226-2022-00051).
\end{acks}

%% file: parts/appendix.tex
\appendix
\section{The Effectiveness of Heterogeneous Dividing Pattern}
Since generating heterogeneous dividing pattern does not require any modification on ultimate training and model architecture, we combine this procedure with other four methods to verify the effectiveness of heterogeneous domain labels. These methods (GroupDRO \cite{Sagawa2020GroupDRO}, CORAL \cite{CORAL}, MMD \cite{li2018domain}, MTL \cite{blanchard2021mtl}) need to distinguish different source domains during their training procedures. We generate new heterogeneous domain labels by the first stage of HTCL. Then we apply these labels to those DG methods and don't change their original backbones. The comparison is shown in Table \ref{tab: plug in}. By applying the procedure, all methods' performances are improved. The gain comes to at least 2\%, which shows the huge potential brought by generating heterogeneous dividing pattern.
\input{tables/plug_in_comparison.tex}

\section{Sensitivity Analysis}
\begin{figure*}[h]
  \centering
  \includegraphics[width=\linewidth]{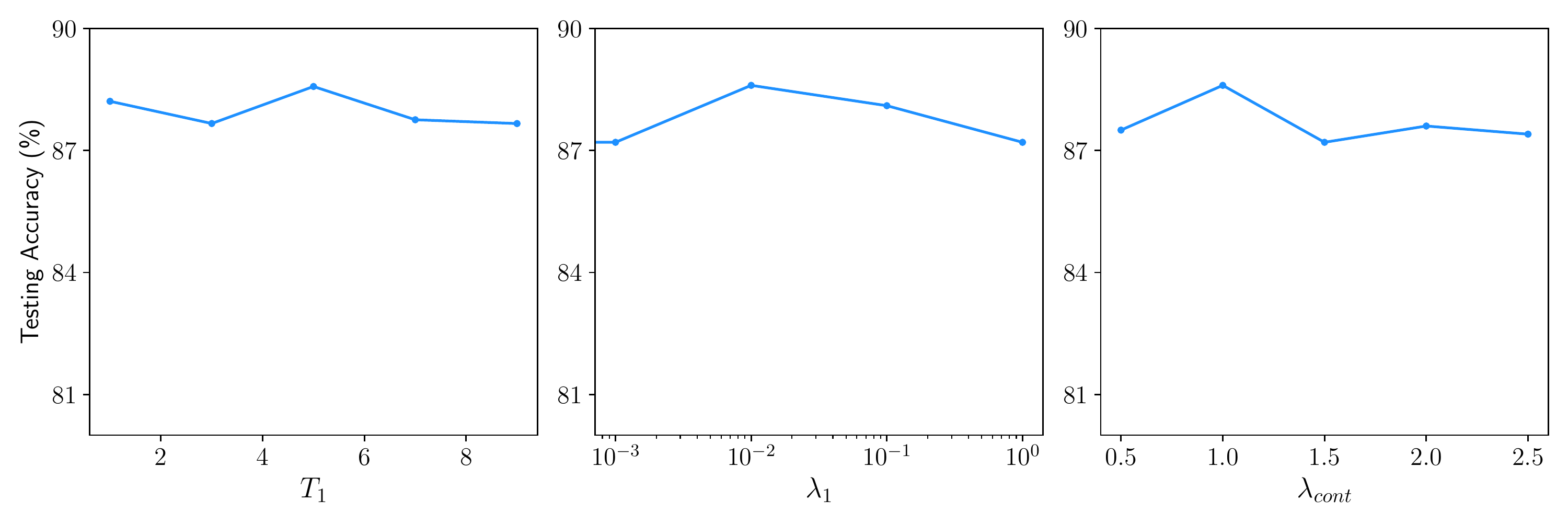}
  \caption{Results of the sensitivity analysis with respect to different hyper-parameters.}\label{fig: sensitive}
\end{figure*}

In this subsection, we study the model sensitivity with respect to the hyper-parameters referred in Algorithm~(\ref{pseudo code}). Table \ref{tab: sensitive} demonstrates the comparison results. The changes of all hyper-parameters don't affect the performances too much and they can all surpass current baselines. As for $\lambda_{mmd}$ and $T_2$ in Algorithm~(\ref{pseudo code}), we set them to fixed 0.5 and 5000 (iterations) respectively as DomainBed \cite{gulrajani2021in} pre-defined to achieve a fair comparison. The visualization for the sensitivity analysis is in Figure~\ref{fig: sensitive}. To demonstrate the results of $\lambda_1$ intuitively, we take the logarithm of its selected values and then map them on the x-axis. It can be seen that our method is robust to the value change of all hyper-parameters.

\begin{table}[h]
\caption{The sensitivity analysis for $T_1$, $\lambda_1$ and $\lambda_{cont}$ on PACS \cite{PACS} dataset.}
\label{tab: sensitive}
\vspace{-8pt}
\begin{tabular}{l|ccccc} 
\toprule
$T_1$ & 1 & 3 & 5 (default) & 7 & 9  \\ 
Test Acc. (\%)  & 88.2 & 87.7 & 88.6 & 87.8 & 88.5 \\
\midrule
$\lambda_1$ & 0 & 1e-3 & 1e-2 (default) & 1e-1 & 1  \\ 
Test Acc. (\%)   & 87.7 & 87.3 & 88.6 & 88.1 & 87.7 \\ 
\midrule
$\lambda_{cont}$ & 0.5 & 1.0 (default) & 1.5 & 2.0 & 2.5  \\ 
Test Acc. (\%)       & 87.5 & 88.6 & 87.2 & 87.6 & 87.4 \\ 
\bottomrule
\end{tabular}
\end{table}

\section{Further Discussion on the Main Results}
Here we further analyze the main results in Table~\ref{table: full_table}. As Table~\ref{table: full_table} illustrates, HTCL doesn't show the superior performance on VLCS \cite{VLCS} dataset. We think it is the original data heterogeneity that limits the performance of our method in specific datasets. Different from other three datasets, data of each domain of VLCS are collected from a specific dataset, which indirectly increases the original heterogeneity of the whole dataset. 
In the first stage of HTCL, we aim to replace original dividing pattern with our newly generated heterogeneous pattern. However, when the given pattern is already heterogeneous enough like VLCS\cite{VLCS}, 
The performance gain brought by heterogeneous dividing pattern generation will naturally be reduced. Therefore, considering the influence of original data heterogeneity before applying our method may be our future improvement.

%% file: tables/plug_in_comparison.tex
\begin{table}[h]
\caption{The results of adding our newly generated dividing pattern to common DG methods on PACS \cite{PACS} dataset. The heterogeneous domain labels are generated by HTCL's first stage. We then apply them on common DG methods. The gain of accuracies is reported in the last column.}
\label{tab: plug in}
\begin{tabular}{lccc} 
\toprule
                & Original &  With new domain labels & $\Delta$ \\
\midrule
GroupDRO \cite{Sagawa2020GroupDRO}      & 84.4 \scriptsize$\pm0.8$ & 88.1 \scriptsize$\pm0.2$  & +3.7 \\
CORAL \cite{CORAL}                      & 86.2 \scriptsize$\pm0.3$ & 88.3 \scriptsize$\pm0.4$ & +2.1 \\
MMD \cite{li2018domain}                 & 84.7 \scriptsize$\pm0.5$ & 87.4 \scriptsize$\pm0.1$ & +2.7 \\
MTL \cite{blanchard2021mtl}             & 84.6 \scriptsize$\pm0.5$ & 88.4 \scriptsize$\pm0.4$ & +3.8 \\
\bottomrule
\end{tabular}
\end{table}